\pdfoutput=1
\documentclass[sigconf]{acmart}
\AtBeginDocument{%
  }

\copyrightyear{2026}
\acmYear{2026}
\setcopyright{cc}
\setcctype{by}
\acmConference[IntRob '26]{Intelligent Robotics FAIR 2026}{June 18--19, 2026}{Budapest, Hungary}
\acmBooktitle{Intelligent Robotics FAIR 2026 (IntRob '26), June 18--19, 2026, Budapest, Hungary}
\acmDOI{10.1145/3831600.3831618}
\acmISBN{979-8-4007-2435-0/26/06}

\usepackage{float}
\usepackage{multirow}

\begin{document}

\title{Neural Networks for Temporal Pattern Recognition and Dynamic Arm Gesture Speed Estimation for Robot Control}

\author{Milán Zsolt Bagladi}
\orcid{0009-0003-5362-7747}
\email{hhpw8b@inf.elte.hu}
\affiliation{%
  \department{Department of Artificial Intelligence}
  \institution{ELTE Eötvös Loránd University}
  \city{Budapest}
  \country{Hungary}
}

\author{László Gulyás}
\orcid{0000-0002-6367-6695}
\email{lgulyas@inf.elte.hu}
\affiliation{%
  \department{Department of Artificial Intelligence}
  \institution{ELTE Eötvös Loránd University}
  \city{Budapest}
  \country{Hungary}
}

\begin{abstract}
Deploying intelligent robotic systems that interact with humans through gestures requires neural networks capable of recognizing diverse temporal patterns. We present a systematic benchmark of ten abstract sequential tasks---five permutation-invariant (set) and five order-dependent (sequence) problems---evaluated across eighteen neural network architectures spanning recurrent, convolutional, attention-based, and set-function families. Beyond the core architecture--task grid, we explore numerous preprocessing and target-variable transformations, yielding more than 250 distinct experimental configurations. All variants are trained and tested under strictly identical conditions (fixed random seeds, shared hyperparameters, shared data splits) to ensure fair and reproducible comparison. Ranking across all ten tasks reveals four consistently top-performing architectures---BiGRU, TCN, Conv1D, and GRUReLU---all compact enough for real-time deployment (under 2\,000 parameters in the benchmark setting). Based on this ranking, we apply three architecturally diverse top models (BiGRU, TCN, and GRUReLU) to a practical robotics problem: estimating the execution speed of dynamic arm gestures from skeletal keypoint sequences. Three speed interpretations (peak count, period time, and mean spike spacing) are evaluated on a custom dataset of eight traffic-related gesture classes comprising 256\,710 frames recorded via OpenPose. The best configuration achieves a mean absolute error of 0.198 on the peak-count interpretation, corresponding to roughly 5\% relative error, while the period-time interpretation reaches approximately 4\% relative error, and the mean spike spacing interpretation approximately 8\% relative error. These results demonstrate that neural networks can reliably estimate gesture speed from skeletal data, opening a path toward speed-aware gesture-controlled robotic systems.
\end{abstract}


\maketitle

\section{Introduction}
\label{s:introduction}

Human arm gestures are used to convey precise, safety-critical commands in a wide range of real-world scenarios: traffic officers regulate vehicle flow at intersections, railway signalers authorize train movements, and airport ground crew guide aircraft during taxiing and parking. In all these cases, automating the recognition of such gestures can improve safety and operational efficiency. In our prior work~\cite{bagladi2025fast}, we developed a real-time pipeline for recognizing the \emph{type} of dynamic arm gestures using OpenPose keypoint extraction, a $1 \times 1$ normalization scheme, and a GRU-based classifier. That system achieved high accuracy across varying camera angles and gesture speeds, and was successfully demonstrated controlling a TurtleBot~3 mobile robot.

However, recognizing \emph{what} gesture is performed is only part of the problem. In many practical scenarios, the \emph{speed} at which a gesture is executed carries additional meaning. For example, a rapid circular arm motion may signal urgency, while a slow one indicates a routine command. Our prior pipeline was shown to be largely insensitive to speed variations during \emph{classification}---an advantage for type recognition, but a limitation when speed itself must be measured.

The speed measurement problem can be formulated as follows: given a sliding window of skeletal keypoint frames depicting a cyclic gesture, estimate a scalar value that characterizes its execution speed. As shown in~\cite{bagladi2025fast}, this can be approached algorithmically by selecting a characteristic \emph{starting position} (null position) for each gesture, computing the per-frame distance from this position, and measuring the number of frames between consecutive local minima of the resulting distance curve. This yields a discrete speed value (in frames per cycle) that can be directly linked to wall-clock time given the camera's frame rate.

A natural question arises: can neural networks learn to perform such speed estimation directly from the raw distance curves, without manually written rules to detect local minima? More broadly, \emph{which neural network architectures are best suited for recognizing abstract temporal patterns in sequential data?}

To answer this question rigorously, we designed a benchmark of ten abstract tasks that isolate specific pattern-recognition capabilities required for speed estimation and related problems. Five tasks are \emph{set-type} (permutation-invariant), where the order of elements does not affect the target value, and five are \emph{sequence-type}, where temporal order is essential. We evaluate eighteen diverse neural network architectures on all ten tasks under strictly controlled, identical experimental conditions, rank the architectures by aggregated per-task performance, and apply the three top-ranked models to a real-world gesture speed estimation problem using three distinct speed interpretations on a custom dataset of eight traffic-related arm gestures. Our experiments show that neural networks can estimate gesture speed from skeletal data with relative errors ranging from 4\% to 8\% across three speed interpretations, validating the approach for practical robotics deployment.

The main contributions of this paper are:
\begin{itemize}
    \item A systematic benchmark of 18 neural network architectures across 10 abstract sequential tasks (5 permutation-invariant, 5 order-dependent), with over 250 experimental configurations evaluated under strictly controlled, identical conditions.
    \item A rank-based aggregation methodology that enables fair cross-task architecture comparison despite heterogeneous output scales, yielding a reproducible global architecture ranking.
    \item Validation of the benchmark findings on a practical robotics problem: neural speed estimation of dynamic arm gestures from skeletal keypoint data, achieving relative errors at or below 5\% on two of three speed interpretations.
\end{itemize}

The paper begins with a review of related work on gesture recognition and neural sequence modeling (Section~\ref{s:related_work}), then introduces the benchmark task suite that captures the key pattern-recognition capabilities needed for speed estimation (Section~\ref{s:benchmark_tasks}). Section~\ref{s:architectures} presents the eighteen evaluated architectures, spanning recurrent, convolutional, attention-based, and set-function families. The experimental protocol and ranking methodology are detailed in Section~\ref{s:experimental_setup}, followed by the benchmark results and global architecture ranking in Section~\ref{s:benchmark_results}. Building on these findings, Section~\ref{s:application} applies the top-ranked models to real gesture speed estimation, and Section~\ref{s:conclusions} concludes with a discussion of future directions.

\section{Related Work}
\label{s:related_work}

\subsection{Gesture Recognition in Robotics}
Recent studies combine keypoint extraction and deep learning for gesture recognition in traffic and robotics. Bagladi~\cite{bagladi2025fast,bagladi2025artificial} uses OpenPose with neural models; He \emph{et al.}~\cite{HE2020248} combine a modified CPM, hand-crafted spatial features, and LSTM; and Zanfir \emph{et al.}~\cite{6751453} enrich pose descriptors with joint speed and acceleration for low-latency action recognition. Vieira \emph{et al.}~\cite{vieira2017online} use speed-aware action graphs, and Stadelmayer \emph{et al.}~\cite{stadelmayer2024lightweight} pair classification with regression of continuous gesture attributes. Yet none estimates arm-gesture execution \emph{speed} from skeletal keypoint data as a continuous scalar.

\subsection{Neural Architectures for Sequential Data}
The architectures evaluated in this work span several families. Recurrent models such as GRU~\cite{cho2014learningphraserepresentationsusing} and LSTM~\cite{hochreiter1997long} process sequences element by element and maintain hidden states that capture temporal dependencies. Bidirectional variants~\cite{schuster1997bidirectional} (BiGRU, BiLSTM) process the sequence in both directions. Temporal Convolutional Networks (TCN)~\cite{bai2018empirical} use dilated causal convolutions to capture long-range dependencies with efficient parallel computation. Transformer-based models~\cite{vaswani2017attention,wen2023transformers} use self-attention to capture global dependencies. Set-function architectures such as DeepSets~\cite{zaheer2017deep} are specifically designed for permutation-invariant problems. Our benchmark systematically compares representatives of all these families on a controlled set of tasks.

\section{Benchmark Tasks}
\label{s:benchmark_tasks}

We define ten tasks that each test a specific capability of neural networks when processing fixed-length sequences of numbers. Each task takes as input a sequence $\mathbf{x} = (x_1, x_2, \dots, x_N)$ of $N$ real values and requires the model to produce a single scalar output $\hat{y}$. The tasks are divided into two groups based on whether the ordering of elements affects the target value.

\subsection{Set Tasks (Permutation-Invariant)}
\label{ss:set_tasks}

For these five tasks, the target value $y$ depends only on the \emph{multiset} of input values, not their order. These tasks are where order-independent architectures such as DeepSets and AttentionPool have a theoretical advantage, though as we will see, how well a model trains in practice matters just as much as its theoretical design. In all five tasks, the target outputs are uniformly distributed over their respective ranges, ensuring balanced training conditions.

\begin{enumerate}
    \item \textbf{Decision:} Determine whether the binary sequence ($x_i \in \{0, 1\}$) contains at least one element equal to~1.
    \[y = \mathbf{1}[\exists\, i : x_i = 1], \quad y \in \{0, 1\}.\]
    This binary classification task is the simplest in the benchmark and serves as a sanity check.

    \item \textbf{Counting:} Count the number of elements equal to~1 in a binary sequence ($x_i \in \{0, 1\}$).
    \[y = \sum_{i=1}^{N} \mathbf{1}[x_i = 1], \quad y \in \{0, 1, \dots, N\}.\]
    All possible counts are represented uniformly in the training data.

    \item \textbf{Variance:} Estimate the variance of the input sequence. Elements are drawn from $\text{Uniform}(\mu - w,\; \mu + w)$, giving $\text{Var}(\mathbf{x}) = w^2/3$.
    \[y = \frac{1}{N}\sum_{i=1}^{N}(x_i - \bar{x})^2, \quad y \in [0, 0.25].\]

    \item \textbf{Mean:} Estimate the arithmetic mean of elements drawn from $[0, 1]$.
    \[y = \frac{1}{N}\sum_{i=1}^{N} x_i, \quad y \in [0, 1].\]

    \item \textbf{Maximum:} Estimate the maximum value of the sequence.
    \[y = \max(x_1, \dots, x_N), \quad y \in [0, 1].\]
\end{enumerate}

\subsection{Sequence Tasks (Order-Dependent)}
\label{ss:sequence_tasks}

For these five tasks, the target value depends on the \emph{ordering} of the input elements. The temporal structure of the input carries essential information, placing models with explicit sequence-processing mechanisms---such as recurrent networks and causal convolutions---at a potential advantage.

\begin{enumerate}
    \setcounter{enumi}{5}
    \item \textbf{Speed:} Estimate the mean absolute consecutive difference of a sequence with values in $[0, 1]$.
    \[y = \frac{1}{N-1}\sum_{i=1}^{N-1} |x_{i+1} - x_i|, \quad y \in [0, 0.5].\]

    \item \textbf{Peak Count:} Count the number of local maxima in the sequence. A point $x_i$ is a peak if $x_{i-1} < x_i > x_{i+1}$. Sequences are generated as AR(1) processes with varying autocorrelation to control peak density.
    \[y = \bigl|\{i : x_{i-1} < x_i > x_{i+1}\}\bigr|, \quad y \in [0, N/2].\]

    \item \textbf{Period Time:} Estimate the number of complete cycles of a sinusoidal signal $x_i = \sin(2\pi f \cdot i/N)$, where $f \in [1, 11]$ is the cycle count within the $N$-sample window.
    \[y = f, \quad y \in [1, 11].\]

    \item \textbf{Spike Spacing:} Estimate the mean temporal distance between spike events in a sparse signal. The sequence consists of a baseline ($\approx 0$) with 2--6~spikes ($\approx 1$) at random positions.
    \[y = \frac{1}{K-1}\sum_{k=1}^{K-1}(p_{k+1} - p_k), \quad y \in [3, 30] \text{ frames},\]
    where $p_1, \dots, p_K$ are the spike positions.

    \item \textbf{Threshold Crossings:} Count the number of upward crossings of the threshold $0.5$.
    \[y = \sum_{i=1}^{N-1} \mathbf{1}[x_i < 0.5 \;\land\; x_{i+1} \geq 0.5].\]
\end{enumerate}

Tasks~7, 8, and~9 (Peak Count, Period Time, and Spike Spacing) were specifically chosen because they correspond to three natural interpretations of gesture execution speed (see Section~\ref{s:application}): counting how many complete gesture cycles occur in a window, measuring the duration of a single cycle, and computing the average inter-cycle distance. The remaining tasks complement these by testing the underlying skills needed---detecting the presence of events (Task~1), counting elements (Task~2), estimating statistical summaries (Tasks~3--5), measuring rate of change (Task~6), and detecting transitions (Task~10).

\section{Neural Network Architectures}
\label{s:architectures}

We evaluate eighteen architectures spanning five broad families. All models map an input sequence $\mathbf{x} \in \mathbb{R}^N$ to a single scalar output $\hat{y} \in \mathbb{R}$ (binary classification for Task~1). Each architecture is described below; Tables~\ref{tab:set_results} and~\ref{tab:seq_results} list the corresponding parameter counts.

\subsection{Recurrent Models}
Six architectures process the sequence element by element, maintaining a hidden state that captures temporal dependencies.
\textbf{GRU}~\cite{cho2014learningphraserepresentationsusing} is a Gated Recurrent Unit whose final hidden state is mapped to the output via a linear layer; we test basic, higher-hidden-dimension, and linear-activation variants.
\textbf{GRUReLU} applies LayerNorm~\cite{ba2016layernormalization} followed by ReLU to the GRU's final hidden state before the linear output layer, replacing the default linear projection and avoiding output saturation.
\textbf{BiGRU}~\cite{schuster1997bidirectional} is a bidirectional GRU that concatenates forward and backward hidden states, followed by LayerNorm and a linear output.
\textbf{LSTM}~\cite{hochreiter1997long} is the classic Long Short-Term Memory network, tested in the same three variants as GRU.
\textbf{BiLSTM} is the bidirectional analogue of BiGRU using LSTM cells.
\textbf{SkipGRU} augments the GRU with skip connections that add the input directly to the hidden state.

\subsection{Convolutional Models}
Two architectures use 1D convolutions to capture local and long-range patterns.
\textbf{Conv1D} applies parallel convolution branches with kernel sizes 1, 3, and 5, followed by global average and max pooling.
\textbf{TCN}~\cite{bai2018empirical} (Temporal Convolutional Network) uses dilated causal convolutions (dilation factors 1, 2, 4) with ReLU, global pooling, and auxiliary features (input mean, $\log N$). Conv1D can be viewed as a simplified, non-causal variant of TCN.

\subsection{Attention and Transformer Models}
Three architectures use attention-based aggregation.
\textbf{Transformer} is a permutation-invariant model built from an element-wise MLP~($\varphi$), multi-aggregation pooling (mean, max, min, std, sum/$N$), and a readout MLP~($\rho$). Unlike standard Transformer architectures, this variant employs \emph{no self-attention} and no positional encoding; its name reflects the aggregation-based readout style it shares with certain Transformer decoder modules. It is functionally a learnable extension of DeepSets with richer pooling statistics.
\textbf{TransformerPosCoding} is identical but prepends sinusoidal positional encoding to each input element, breaking permutation invariance and making it order-aware.
\textbf{AttentionPool} learns attention weights via a scoring network ($\text{Linear} \to \tanh \to \text{Linear} \to \text{softmax}$) and produces attended sum, max, mean, and sum aggregations.

\subsection{Set-Function Models}
Six architectures are designed around permutation-invariant primitives.
\textbf{DeepSets}~\cite{zaheer2017deep} applies an element-wise MLP~($\varphi$), mean/max/sum pooling, and a readout MLP~($\rho$); it is theoretically universal for permutation-invariant functions.
\textbf{Histogram} learns soft bin assignments via Gaussian kernels with trainable centers, producing a fixed-size histogram processed by an MLP.
\textbf{Quantile} computes 21 quantile points (0\%, 5\%, \dots, 100\%) of the input, yielding a fixed-size summary processed by an MLP.
\textbf{LogSumExp} approximates smooth maximum/minimum via the log-sum-exp operator with learnable temperature parameters.
\textbf{PowerMean} computes generalized power means $M_p(\mathbf{x}) = \left(\frac{1}{N}\sum x_i^p\right)^{1/p}$ with learnable exponents~$p$.
\textbf{ElemAgg} applies element-wise learnable thresholds and aggregates six features (sum, mean, max of both transformed and raw values).

\subsection{Baseline}
\textbf{NN} is a simple feedforward network $\text{Linear}(N \to 1)$ that treats the sequence as a flat input vector. It is \emph{not} length-invariant and serves as a na\"ive baseline.

\section{Experimental Setup}
\label{s:experimental_setup}

\subsection{Data Generation}

For each task, training and test data are generated synthetically with full control over the target distribution. A key design principle is that target values are \emph{uniformly distributed} across the output range, preventing bias toward any particular output region. Sequences of fixed length $N = 50$ are generated with controlled parameters specific to each task, matching the sliding window used in the gesture speed estimation application (Section~\ref{s:application}).

Multiple dataset size configurations were tested: $N_{\text{train}}{:}N_{\text{test}} \in \{50{:}50,\; 250{:}250,\; 1000{:}1000,\; 50{:}1000,\; 50{:}10000\}$, where $N_{\text{train}}$ and $N_{\text{test}}$ denote the number of training and test samples, respectively. Symmetric configurations ($N_{\text{train}} = N_{\text{test}}$) evaluate standard performance under matched conditions, while asymmetric ones ($50{:}1000$, $50{:}10000$) test how well models generalize to a more densely sampled output range than they were trained on. The primary evaluation uses the $1000{:}1000$ configuration. The asymmetric configurations confirmed consistent relative architecture rankings and are therefore not reported separately.

\subsection{Fair Comparison Protocol}

To ensure that observed performance differences reflect genuine architectural strengths rather than other influences, every experiment follows a strict fair-comparison protocol. The core grid comprises $18$ architectures $\times$ $10$ tasks, but for several architectures---notably GRU, LSTM, and SkipGRU---we additionally explore multiple preprocessing and target-variable transformations such as linear scaling, density-based encoding, hidden-dimension variants, and their combinations. This yields more than 250 distinct experimental configurations, documented in over 500 individual training and evaluation notebooks.

Across all configurations, three invariants are maintained. First, fixed random seeds guarantee that every architecture receives exactly the same data splits for each task, so that data variability cannot affect the comparison. Second, all models share identical hyperparameters: Adam optimizer with a learning rate of 0.01, batch size of 32, a maximum of 200 training epochs, and early stopping with a patience of 5 epochs. Third, evaluation metrics are uniform: Task~1 is assessed by accuracy (equivalently, error rate $= 1 - \text{acc}$), while Tasks~2--10 are assessed by both Mean Absolute Error (MAE) and Mean Squared Error (MSE).

\subsection{Ranking Methodology}

Because each task operates on a different output scale and unit, raw MAE and MSE values are not directly comparable across tasks. A low MAE on one task may represent excellent performance while the same numerical value on another task may be poor. To obtain a fair cross-task comparison, we therefore adopt a rank-based aggregation strategy. For each task, all 18~architectures are sorted by their primary metric (MAE for Tasks~2--10, accuracy for Task~1) and assigned \emph{dense ranks}: metric values are rounded to three decimal places before comparison, rank~1 is best, ties (equal rounded values) receive the same rank, and the next distinct value receives rank~$+1$ (no gaps). Architectures that fail to produce valid output on a task (e.g., diverged training) receive one rank below the last valid result on that task.

The \emph{global rank sum} for each architecture is the sum of its per-task ranks: $\Sigma_{\text{rank}} = \sum_{t=1}^{10} r_t,$
where $r_t$ is the architecture's dense rank on task~$t$. Lower $\Sigma_{\text{rank}}$ values indicate more consistently strong performance across the full task suite. 

\section{Benchmark Results}
\label{s:benchmark_results}

\subsection{Set Tasks (1--5)}

Table~\ref{tab:set_results} summarizes the performance of all architectures on the five set tasks using the $1000{:}1000$ configuration. The metric shown is accuracy for Task~1 and MAE for Tasks~2--5.

\begin{table*}[t]
    \centering
    \caption{Performance on set tasks ($1000{:}1000$ configuration). Task~1: accuracy ($\uparrow$); Tasks~2--5: MAE ($\downarrow$). Values rounded to 3 decimal places. \textbf{Bold} = rank~1. Superscript = per-task dense rank (ties share the same rank). -- = convergence failure (ranked last).}
    \label{tab:set_results}
    \small
    \begin{tabular}{l|c|c|c|c|c|c}
        \hline
        \textbf{Architecture} & \textbf{Params} & \textbf{1-Decision} & \textbf{2-Counting} & \textbf{3-Variance} & \textbf{4-Mean} & \textbf{5-Max} \\
        \hline
        GRU          & 14     & \textbf{1.000} & 3.428$^{11}$  & \textbf{0.004} & 0.001$^{2}$            & 0.006$^{4}$ \\
        GRUReLU      & 963    & 0.994$^{2}$    & 1.226$^{9}$   & \textbf{0.004} & 0.002$^{3}$            & 0.011$^{9}$ \\
        BiGRU        & 1,921  & \textbf{1.000} & 2.020$^{10}$  & \textbf{0.004} & 0.001$^{2}$            & 0.007$^{5}$ \\
        LSTM         & 16     & 0.960$^{5}$    & 12.726$^{14}$ & 0.025$^{4}$    & 0.012$^{7}$            & 0.020$^{11}$ \\
        BiLSTM       & 2,529  & 0.988$^{4}$    & 14.126$^{15}$ & 0.012$^{3}$    & 0.015$^{9}$            & 0.033$^{13}$ \\
        SkipGRU      & 14     & \textbf{1.000} & 19.444$^{16}$ & 0.062$^{9}$    & 0.014$^{8}$            & 0.044$^{14}$ \\
        \hline
        Conv1D       & 292    & \textbf{1.000} & 0.108$^{5}$   & 0.012$^{3}$    & 0.004$^{4}$            & 0.009$^{7}$ \\
        TCN          & 1,667  & \textbf{1.000} & 0.502$^{7}$   & 0.005$^{2}$    & 0.001$^{2}$            & 0.009$^{7}$ \\
        \hline
        Transformer  & 24,961 & \textbf{1.000} & 0.006$^{3}$   & 0.063$^{10}$   & 0.004$^{4}$            & \textbf{0.003} \\
        TransfPosCod & 29,665 & 0.500$^{6}$    & 3.890$^{12}$  & 0.063$^{10}$   & 0.250$^{13}$           & 0.250$^{15}$ \\
        AttentionPool& 30     & \textbf{1.000} & 0.670$^{8}$   & 0.225$^{13}$   & 0.218$^{12}$           & 0.029$^{12}$ \\
        \hline
        DeepSets     & 16,705 & 0.500$^{6}$    & \textbf{0.000}& 0.044$^{7}$    & 0.005$^{5}$            & 0.013$^{10}$ \\
        Histogram    & 406    & 0.994$^{2}$    & 0.022$^{4}$   & 0.042$^{6}$    & 0.001$^{2}$            & 0.009$^{7}$ \\
        Quantile     & 385    & \textbf{1.000} & 12.537$^{13}$ & 0.012$^{3}$    & \textbf{0.000}         & 0.005$^{3}$ \\
        LogSumExp    & 12     & \textbf{1.000} & \textbf{0.000}& 0.060$^{8}$    & 0.010$^{6}$            & 0.010$^{8}$ \\
        PowerMean    & 11     & 0.500$^{6}$    & --            & 0.129$^{12}$   & \textbf{0.000}         & 0.004$^{2}$ \\
        ElemAgg      & 9      & \textbf{1.000} & 0.001$^{2}$   & 0.028$^{5}$    & 0.026$^{11}$           & 0.008$^{6}$ \\
        \hline
        NN           & 1,001  & 0.992$^{3}$    & 0.266$^{6}$   & 0.069$^{11}$   & 0.024$^{10}$           & 0.010$^{8}$ \\
        \hline
    \end{tabular}
\end{table*}

Key observations on set tasks:
\begin{itemize}
    \item \textbf{Task~1 (Decision)} is solved perfectly by 10 of 18 architectures, confirming it as an easy baseline. GRUReLU and Histogram are tied at rank~2 (both 0.994).
    \item \textbf{Task~2 (Counting)} creates the sharpest spread: DeepSets and LogSumExp are tied at rank~1 with MAE~$=$~0.000, followed by ElemAgg (0.001, rank~2) and Transformer (0.006, rank~3). Recurrent models (LSTM, BiLSTM, SkipGRU) struggle with MAE~$>$~12.
    \item \textbf{Task~3 (Variance)} is tied at rank~1 among GRU, GRUReLU, and BiGRU (all 0.004), followed by TCN (0.005).
    \item \textbf{Task~4 (Mean)} has a two-way rank-1 tie: Quantile and PowerMean (both 0.000), and a four-way rank-2 tie (GRU, BiGRU, TCN, Histogram, all 0.001).
    \item \textbf{Task~5 (Max)} is best solved by Transformer (0.003), followed by PowerMean (0.004) and Quantile (0.005).
    \item Notably, the order-independent architectures (DeepSets, Histogram, Quantile, LogSumExp, AttentionPool) do not uniformly outperform sequence-oriented ones on these set tasks, confirming that training behavior matters as much as theoretical design.
\end{itemize}

\subsection{Sequence Tasks (6--10)}

Table~\ref{tab:seq_results} shows results for the five sequence tasks.

\begin{table*}[t]
    \centering
    \caption{Performance on sequence tasks ($1000{:}1000$ configuration, MAE $\downarrow$). Values rounded to 3 decimal places. \textbf{Bold} = rank~1. Superscript = per-task dense rank (ties share the same rank). -- = convergence failure (ranked last).}
    \label{tab:seq_results}
    \small
    \begin{tabular}{l|c|c|c|c|c|c}
        \hline
        \textbf{Architecture} & \textbf{Params} & \textbf{6-Speed} & \textbf{7-Peak} & \textbf{8-Period} & \textbf{9-SpikeSp} & \textbf{10-Threshold} \\
        \hline
        GRU          & 14     & \textbf{0.004}  & 225.496$^{18}$ & 2.500$^{13}$  & 6.752$^{2}$  & 10.318$^{13}$ \\
        GRUReLU      & 963    & 0.006$^{3}$     & 123.024$^{11}$ & 0.074$^{4}$   & 6.752$^{2}$  & 6.294$^{2}$ \\
        BiGRU        & 1,921  & 0.008$^{4}$     & 8.023$^{3}$    & 0.031$^{2}$   & 6.752$^{2}$  & \textbf{6.243} \\
        LSTM         & 16     & 0.083$^{9}$     & 225.267$^{17}$ & 2.500$^{13}$  & 6.753$^{3}$  & 133.807$^{16}$ \\
        BiLSTM       & 2,529  & 0.069$^{7}$     & 113.481$^{10}$ & 0.092$^{5}$   & 6.752$^{2}$  & 6.302$^{3}$ \\
        SkipGRU      & 14     & 0.063$^{6}$     & 180.412$^{15}$ & 2.501$^{14}$  & 6.759$^{6}$  & 241.764$^{17}$ \\
        \hline
        Conv1D       & 292    & 0.006$^{3}$     & 18.842$^{5}$   & \textbf{0.030}& 6.754$^{4}$  & 8.516$^{10}$ \\
        TCN          & 1,667  & 0.005$^{2}$     & 8.361$^{4}$    & \textbf{0.030}& \textbf{4.894} & 7.163$^{4}$ \\
        \hline
        Transformer  & 24,961 & 0.073$^{8}$     & 6.794$^{2}$    & 1.923$^{8}$   & 6.753$^{3}$  & 10.346$^{14}$ \\
        TransfPosCod & 29,665 & 0.084$^{10}$    & 180.538$^{16}$ & 2.871$^{15}$  & 6.774$^{7}$  & -- \\
        AttentionPool& 30     & 0.119$^{14}$    & 153.432$^{13}$ & 1.979$^{11}$  & 6.798$^{9}$  & 8.314$^{6}$ \\
        \hline
        DeepSets     & 16,705 & 0.084$^{10}$    & \textbf{5.203} & 2.028$^{12}$  & 6.754$^{4}$  & 8.496$^{9}$ \\
        Histogram    & 406    & 0.084$^{10}$    & 20.789$^{6}$   & 1.801$^{6}$   & 6.753$^{3}$  & 10.424$^{15}$ \\
        Quantile     & 385    & 0.061$^{5}$     & 123.444$^{12}$ & 1.823$^{7}$   & 6.752$^{2}$  & 8.979$^{12}$ \\
        LogSumExp    & 12     & 0.089$^{12}$    & 107.060$^{9}$  & 1.975$^{10}$  & 6.756$^{5}$  & 8.333$^{8}$ \\
        PowerMean    & 11     & 0.087$^{11}$    & 153.812$^{14}$ & 1.973$^{9}$   & 6.803$^{10}$ & 8.329$^{7}$ \\
        ElemAgg      & 9      & 0.089$^{12}$    & 72.846$^{7}$   & 1.975$^{10}$  & 6.775$^{8}$  & 8.657$^{11}$ \\
        \hline
        NN           & 1,001  & 0.094$^{13}$    & 87.444$^{8}$   & 0.034$^{3}$   & 7.615$^{11}$ & 7.268$^{5}$ \\
        \hline
    \end{tabular}
\end{table*}

Key observations on sequence tasks:
\begin{itemize}
    \item \textbf{Task~6 (Speed)} is best solved by GRU (0.004), followed by TCN (0.005), then GRUReLU and Conv1D tied at rank~3 (both 0.006). Many architectures converge to MAE~$\approx 0.084$, indicating they learn only a constant-output baseline.
    \item \textbf{Task~7 (Peak Count)} strongly differentiates architectures: DeepSets (5.203), Transformer (6.794), and BiGRU (8.023) succeed, while most recurrent models fail entirely (MAE~$> 100$).
    \item \textbf{Task~8 (Period Time)} is tied at rank~1 between Conv1D and TCN (both 0.030), with BiGRU (0.031) close behind. This task rewards architectures capable of detecting periodic structure.
    \item \textbf{Task~9 (Spike Spacing)} is dominated by TCN (4.894, rank~1). The remaining 17 models converge to $\approx 6.752$--$6.803$, yielding a large tied block at the baseline mean prediction.
    \item \textbf{Task~10 (Threshold)} is relatively difficult for all architectures, with BiGRU (6.243) and GRUReLU (6.294) achieving the best results.
\end{itemize}

\subsection{Global Ranking}

\begin{figure*}[t]
    \centering
    \includegraphics[width=\textwidth]{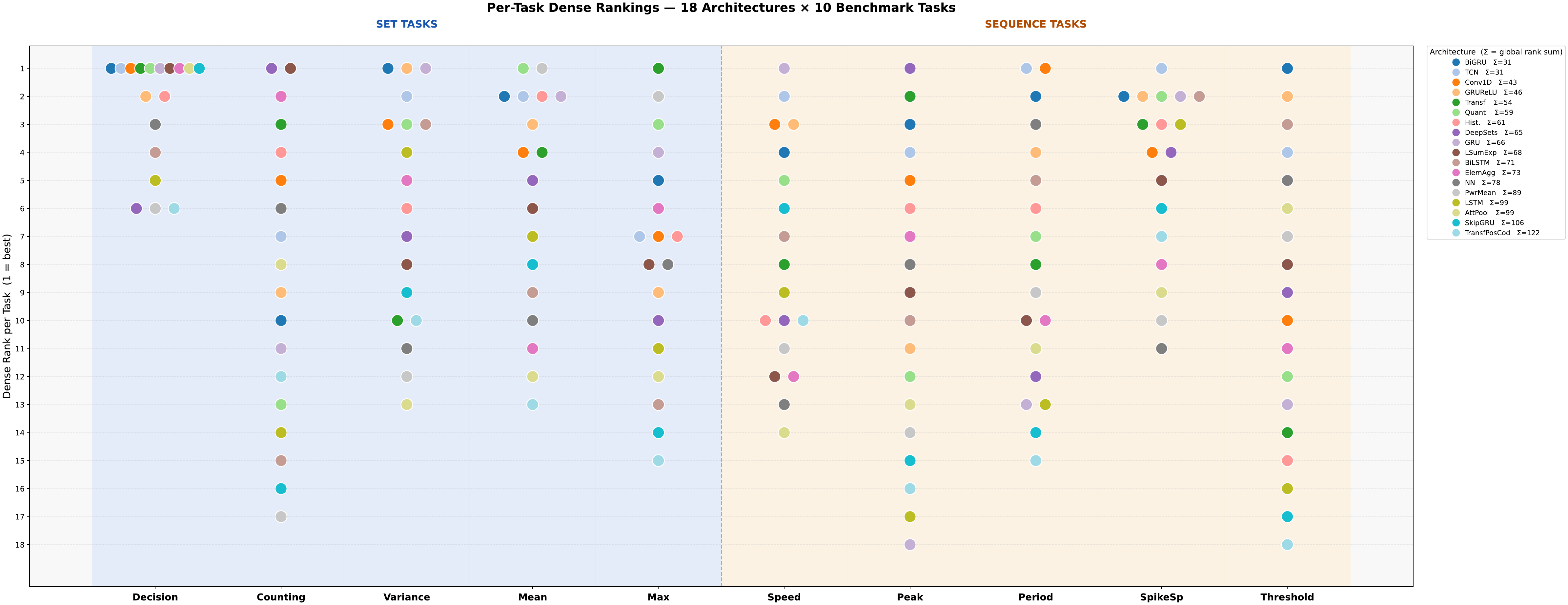}
    \caption{Per-task dense rank of all 18 architectures across the 10 benchmark tasks ($1000{:}1000$ configuration). Each point shows one architecture's rank on a single task; legend entries are ordered by global rank sum ($\Sigma$). Blue background: permutation-invariant set tasks (Tasks~1--5); orange: order-dependent sequence tasks (Tasks~6--10).}
    \Description{Bump chart with one colored series per architecture, showing each model's per-task rank across the ten benchmark tasks.}
    \label{fig:bump_chart}
\end{figure*}

\begin{figure}[!htbp]
    \centering
    \includegraphics[width=\columnwidth]{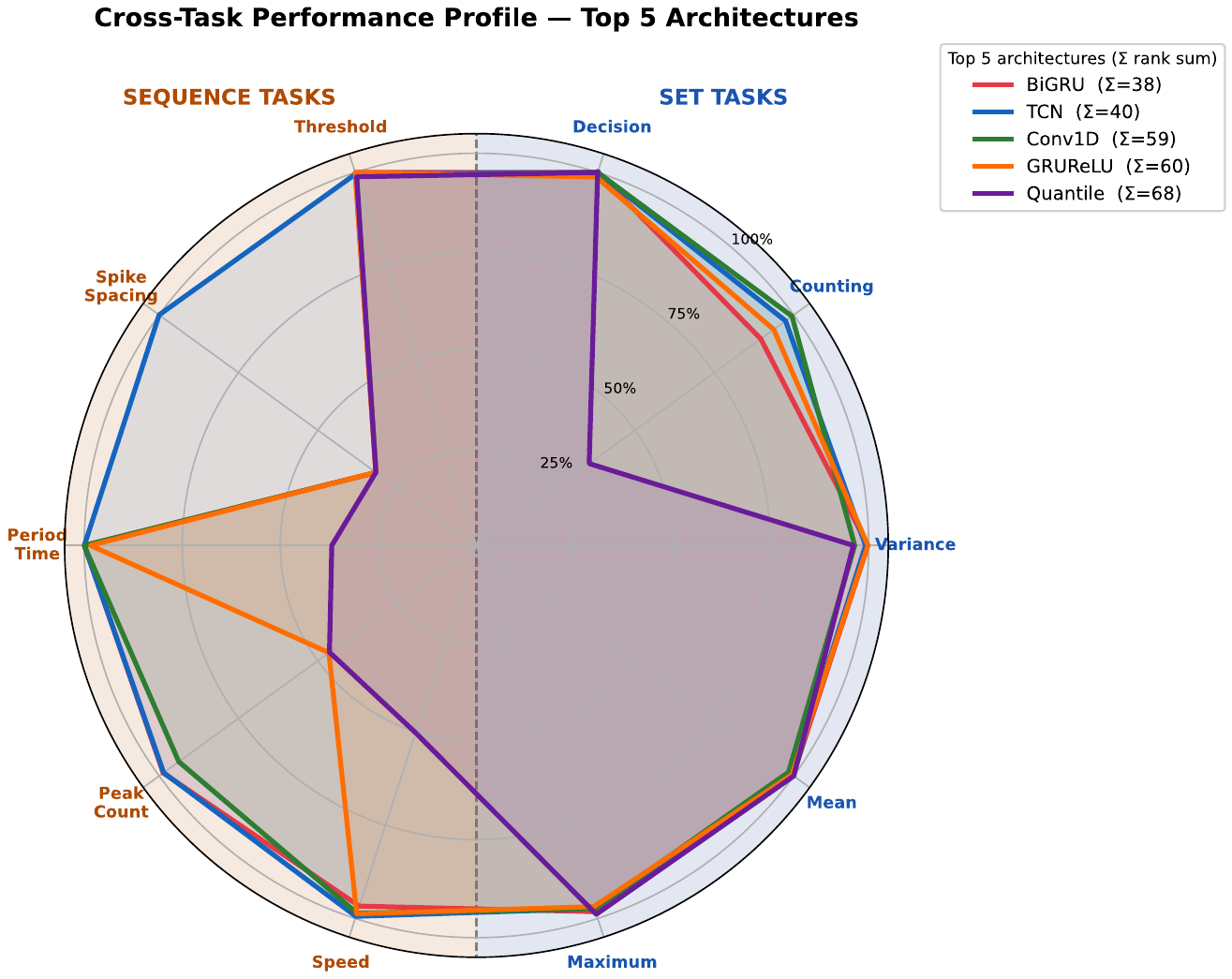}
    \caption{Cross-task performance profiles of the five architectures with the lowest global rank sum ($1000{:}1000$ configuration). Spoke values are min--max normalized across all 18 architectures (1.0 = best, 0.0 = worst). Blue sector: set tasks; orange sector: sequence tasks.}
    \Description{Radar chart showing ten task spokes; each of the five top-ranked architectures is drawn as a colored polygon indicating its normalized score on every task.}
    \label{fig:radar_chart}
\end{figure}

Table~\ref{tab:global_ranking} summarizes the global rank sum $\Sigma_{\text{rank}}$ for all architectures. \textbf{BiGRU} and \textbf{TCN} are jointly ranked first with $\Sigma_{\text{rank}} = 31$, and both lead in sequence tasks ($\Sigma_{\text{SEQ}} = 12$). Conv1D ranks second ($\Sigma_{\text{rank}} = 43$) and GRUReLU third ($46$). Notably, GRU, BiGRU, TCN, and Transformer share the same $\Sigma_{\text{SET}} = 19$, yet differ substantially in $\Sigma_{\text{SEQ}}$, confirming that set-task performance alone does not predict overall quality. AttentionPool and LSTM are tied at rank~14 ($\Sigma_{\text{rank}} = 99$). This pattern is visually captured in Figure~\ref{fig:bump_chart}, while Figure~\ref{fig:radar_chart} highlights the cross-task profile differences among the top-ranked models.

\begin{table}[h]
    \centering
    \caption{All architectures ranked by global rank sum ($\Sigma_{\text{rank}}$, lower is better). $\Sigma_{\text{SET}}$ and $\Sigma_{\text{SEQ}}$ show rank sums for set and sequence tasks, respectively.}
    \label{tab:global_ranking}
    \small
    \begin{tabular}{c|l|c|c|c|c}
        \hline
        \textbf{\#} & \textbf{Model} & \textbf{Params} & $\boldsymbol{\Sigma_{\text{rank}}}$ & $\boldsymbol{\Sigma_{\text{SET}}}$ & $\boldsymbol{\Sigma_{\text{SEQ}}}$ \\
        \hline
        1 & BiGRU        & 1,921  & 31  & 19 & 12 \\
        1 & TCN          & 1,667  & 31  & 19 & 12 \\
        2 & Conv1D       & 292    & 43  & 20 & 23 \\
        3 & GRUReLU      & 963    & 46  & 24 & 22 \\
        4 & Transformer  & 24,961 & 54  & 19 & 35 \\
        5 & Quantile     & 385    & 59  & 21 & 38 \\
        6 & Histogram    & 406    & 61  & 21 & 40 \\
        7 & DeepSets     & 16,705 & 65  & 29 & 36 \\
        8 & GRU          & 14     & 66  & 19 & 47 \\
        9 & LogSumExp    & 12     & 68  & 24 & 44 \\
       10 & BiLSTM       & 2,529  & 71  & 44 & 27 \\
       11 & ElemAgg      & 9      & 73  & 25 & 48 \\
       12 & NN           & 1,001  & 78  & 38 & 40 \\
       13 & PowerMean    & 11     & 89  & 38 & 51 \\
       14 & AttentionPool& 30     & 99  & 46 & 53 \\
       14 & LSTM         & 16     & 99  & 41 & 58 \\
       15 & SkipGRU      & 14     & 106 & 48 & 58 \\
       16 & TransfPosCod & 29,665 & 122 & 56 & 66 \\
        \hline
    \end{tabular}
\end{table}

For the real-world application we select three architectures from the top of the global ranking: \textbf{BiGRU} (rank~1), \textbf{TCN} (rank~1), and \textbf{GRUReLU} (rank~3). The second-ranked Conv1D is skipped because it is architecturally a simplified, non-causal variant of TCN; both rely on 1D convolutions with pooling, so including both would add little architectural diversity. GRUReLU, by contrast, belongs to the recurrent family but differs substantially from BiGRU: it uses a unidirectional GRU with LayerNorm and ReLU rather than a bidirectional pass, giving it a different way of processing the input. The resulting trio---one bidirectional recurrent model, one causal convolutional model, and one unidirectional recurrent model---provides broad architectural coverage while remaining compact (under 2\,000 parameters in the benchmark configuration).

\section{Application: Gesture Speed Estimation}
\label{s:application}

Having identified the top-performing architectures on abstract tasks, we now apply them to a concrete robotics-relevant problem: estimating the execution speed of dynamic arm gestures from skeletal keypoint data.

\subsection{Motivation}

In our prior work~\cite{bagladi2025fast}, we developed a real-time pipeline for dynamic arm gesture recognition that classifies the \emph{type} of gesture (e.g., hand circles, hand waves, call-to-pass). That system uses OpenPose~\cite{cao2019openposerealtimemultiperson2d} for keypoint extraction, $1 \times 1$ normalization, and a GRU-based classifier. The system was shown to be robust to variations in both viewing angle and gesture speed.

However, in many robotic applications, the \emph{speed} at which a gesture is executed carries additional meaning beyond its type---a fast gesture may signal an urgent command, while the same gesture performed slowly indicates a routine one. Our prior pipeline cannot measure this speed; it only classifies gesture type. To address this gap, we propose using neural networks to estimate gesture speed directly from the processed keypoint data.

\subsection{Dataset}

The gesture dataset consists of real recordings of eight dynamic arm gestures captured using a standard camera and processed through the OpenPose BODY-25 model:

\begin{itemize}
    \item \texttt{balraKoroz} (Right Hand Circle Left) -- 18\,253 frames
    \item \texttt{jobbraKoroz} (Right Hand Circle Right) -- 44\,868 frames
    \item \texttt{simaAllas} (Standing Still) -- 2\,783 frames
    \item \texttt{balMeneszt} (Left Hand Wave) -- 40\,081 frames
    \item \texttt{jobbMeneszt} (Right Hand Wave) -- 45\,680 frames
    \item \texttt{felhivasAthaladasra} (Call to Pass) -- 48\,757 frames
    \item \texttt{balkezJobbraKoroz} (Left Hand Circle Right) -- 25\,720 frames
    \item \texttt{balkezBalraKoroz} (Left Hand Circle Left) -- 30\,568 frames
\end{itemize}

All data undergoes the same preprocessing pipeline as described in~\cite{bagladi2025fast}. First, OpenPose BODY-25 keypoints are extracted from each frame. Positional dependence is removed by translating all keypoints so that the neck (BODY-25 keypoint~1) is at the origin. A \emph{$1 \times 1$ normalization} is then applied: the translated keypoints are stretched horizontally and vertically so that the maximum horizontal and vertical distances between any two points become exactly~1, mapping the skeleton into a unit square while preserving its essential spatial structure. From the normalized skeleton, five joint angles are computed: the right elbow (RShoulder--RElbow--RWrist), the left elbow (LShoulder--LElbow--LWrist), the right shoulder (Neck--RShoulder--RElbow), the left shoulder (Neck--LShoulder--LElbow), and the neck-trunk angle (Nose--Neck--MidHip). Each angle is measured over the full $360^\circ$ range and mapped to $[0, 1]$.

A sliding window of 50 frames is used to create input samples. As established in~\cite{bagladi2025fast}, this window length corresponds to approximately 1.67~seconds at 30~FPS, which is sufficient for recognizing a complete gesture cycle and, by extension, for estimating execution speed from the windowed signal.

\subsection{Speed Ground Truth Generation}

\begin{figure}[H]
    \centering
    \includegraphics[width=\columnwidth]{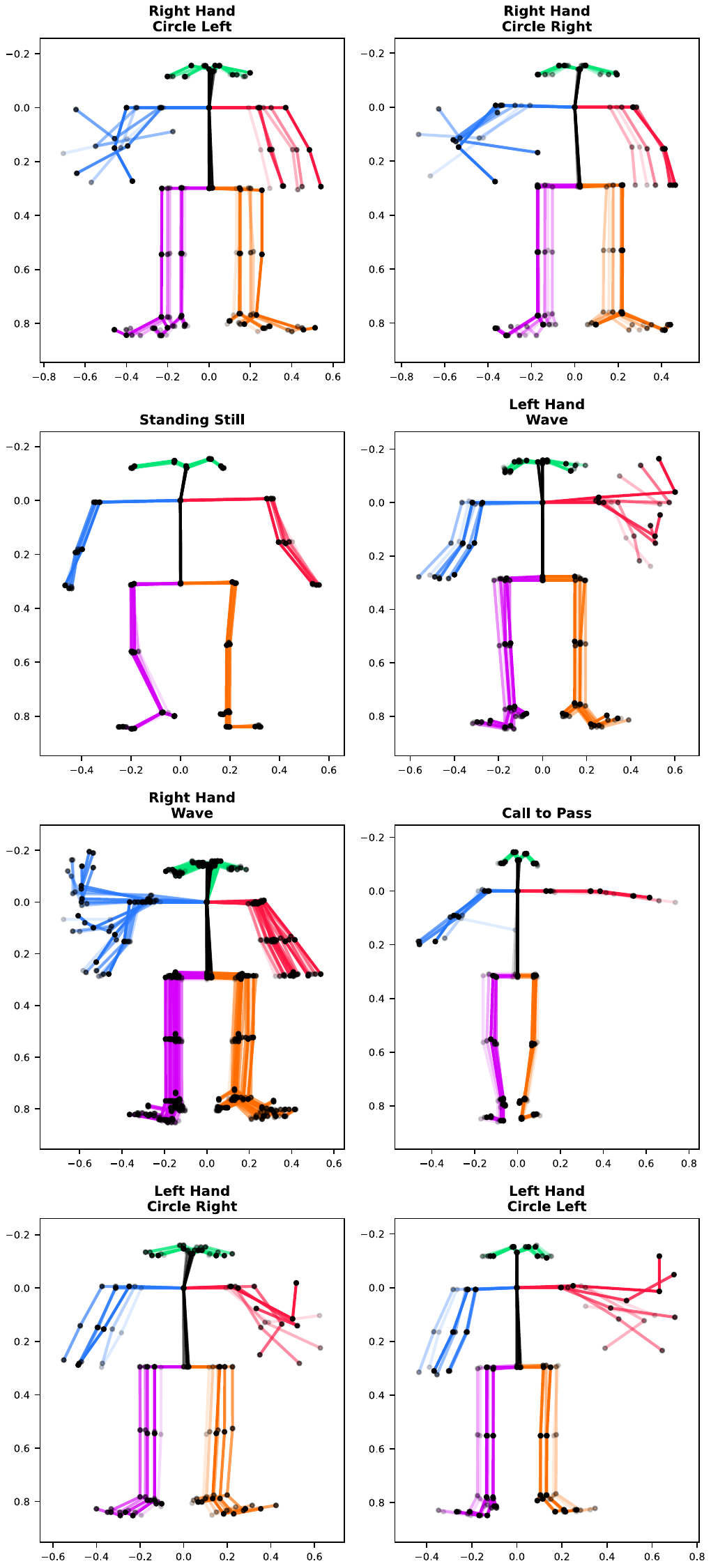}
    \caption{Reference poses for each of the eight gesture types, visualized as BODY-25 skeletal keypoints after $1 \times 1$ normalization. Blue: right arm; red: left arm; black: torso; green: head.}
    \Description{Eight skeleton plots showing the normalized reference position for each gesture type used in the speed estimation application.}
    \label{fig:starting_positions}
\end{figure}

Figure~\ref{fig:starting_positions} depicts the eight gesture types, each shown through their characteristic motion phases. For each gesture, a characteristic \emph{starting position} (also called \emph{null position})---a pose that is periodically revisited during the gesture cycle---is manually identified. The distance from the starting-position skeleton is then computed frame by frame as a weighted Euclidean distance over the nine upper-body keypoints:
\[
d(a,b) = \sqrt{\sum_{i=0}^{8} w_i \left[(a_i^x - b_i^x)^2 + (a_i^y - b_i^y)^2\right]},
\]
where $w_i = 10$ for the four distal arm joints (RElbow, RWrist, LElbow, LWrist; indices~3, 4, 6, 7) and $w_i = 1$ for all others. The elevated weight on elbows and wrists reflects their dominant role in the performed gestures; lower-body keypoints (indices~9--24) are excluded entirely.

\begin{figure*}[htb]
    \centering
    \includegraphics[width=0.6\textwidth]{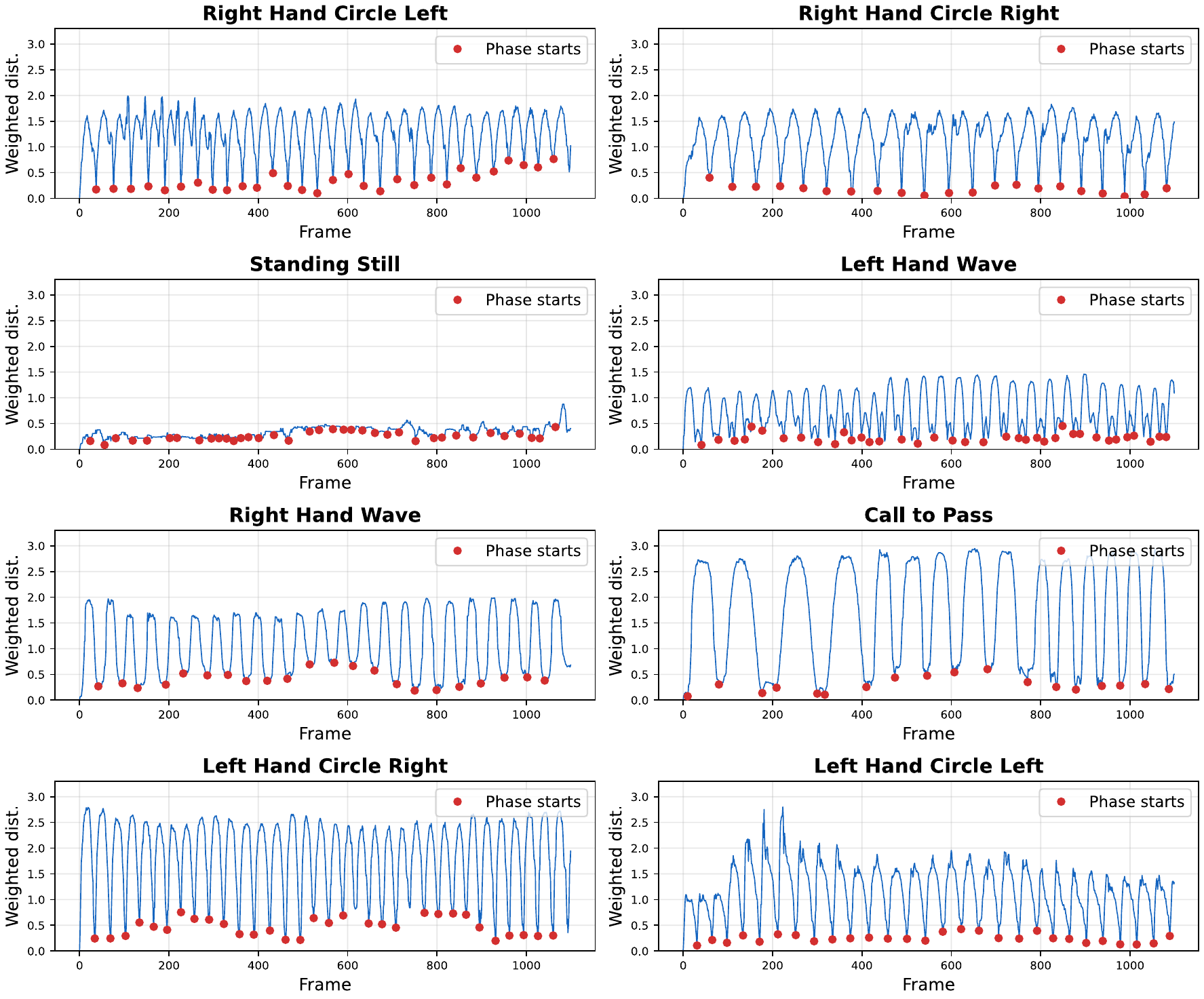}
    \caption{Weighted distance-from-reference-pose curves for example gesture sequences. Red dots mark local minima---moments at which the skeleton is closest to the reference pose. Inter-minima spacing reflects execution speed.}
    \label{fig:distance_curves}
    \Description{Line plots showing the weighted distance from the reference pose over time for different gestures, with local minima marked by red dots.}
\end{figure*}

Computed over the entire sequence, this yields a 1D distance curve (Figure~\ref{fig:distance_curves}) whose local minima correspond to moments when the gesture returns to (or passes closest to) the starting position; their temporal spacing directly encodes the execution speed. This algorithmic approach was established in~\cite{bagladi2025fast} and serves as the labeling mechanism for all three speed interpretations described below.

\subsection{Three Speed Interpretations}

The speed estimation problem admits three distinct interpretations. Each corresponds to one of our benchmark tasks and captures a different---but equally valid---aspect of execution speed from the distance curve described above. This is precisely why Tasks~7, 8, and~9 were included in the benchmark: they abstract the core pattern-recognition capabilities that the application requires. Conversely, the benchmark results guide our choice of models for the application.

In all three cases, the model receives as input the 5~normalized joint angles over 50~frames. The three interpretations differ in what target value $y$ is extracted from the local minima of the distance curve:

\begin{enumerate}
    \item \textbf{Peak Count (Task~7):} The target is the number of local minima (i.e., returns to the starting position) within the 50-frame window:
    \[y = \bigl|\{i : d_{i-1} > d_i < d_{i+1}\}\bigr|,\]
    where $d_i$ is the distance from the starting position at frame~$i$. A higher count indicates faster execution. This interpretation directly counts how many complete gesture cycles fit in the window.

    \item \textbf{Period Time (Task~8):} The target is the number of frames between the first two consecutive local minima, measuring the duration of a single cycle:
    \[y = p_2 - p_1 \quad \text{(in frames)},\]
    where $p_1, p_2$ are the frame indices of the first two minima. A lower value indicates faster execution. Given the camera's frame rate, this can be directly converted to wall-clock time (e.g., at 30~FPS, $y = 30$ corresponds to a 1-second cycle).

    \item \textbf{Spike Spacing (Task~9):} The target is the mean distance (in frames) between all consecutive pairs of local minima in the window:
    \[y = \frac{1}{K-1}\sum_{k=1}^{K-1}(p_{k+1} - p_k),\]
    where $p_1, \dots, p_K$ are the positions of the $K$~local minima. This provides a smoothed, more robust estimate of cycle duration by averaging over all available cycles rather than relying on a single pair.
\end{enumerate}

\subsection{Results}

Table~\ref{tab:application_results} presents the results of applying the three selected architectures to the three speed interpretations on real gesture data. The dataset is split 80\%/20\% into training and test sets using a fixed random seed for reproducibility.

\begin{table}[htbp]
    \centering
    \caption{Application results on real gesture speed estimation. \textbf{Bold} = best per interpretation group: MAE, MSE, Std ($\downarrow$); Params ($\downarrow$ fewest); Bias ($|\cdot|$ closest to zero). Bias is the signed mean error; Std is the error standard deviation.}
    \label{tab:application_results}
    \small
    \begin{tabular}{l|l|r|r|r|r|r}
        \hline
        \textbf{Interpr.} & \textbf{Model} & \textbf{Params} & \textbf{MAE} & \textbf{MSE} & \textbf{Bias} & \textbf{Std.} \\
        \hline
        \multirow{3}{*}{7-Peak}
            & BiGRU   & 27,649 & \textbf{0.198} & \textbf{0.097} & \textbf{$+$0.009} & \textbf{0.311} \\
            & GRUReLU & 13,825 & 0.214 & 0.113 & \textbf{$-$0.009} & 0.336 \\
            & TCN     & \textbf{6,785}  & 0.272 & 0.141 & $+$0.065 & 0.370 \\
        \hline
        \multirow{3}{*}{8-Period}
            & GRUReLU & 13,057 & \textbf{1.320} & \textbf{5.942} & \textbf{$-$0.220} & \textbf{2.428} \\
            & BiGRU   & 26,113 & 1.499 & 7.157 & $+$0.478 & 2.632 \\
            & TCN     & \textbf{6,401}  & 2.882 & 16.756 & $+$0.775 & 4.019 \\
        \hline
        \multirow{3}{*}{9-SpikeSp}
            & BiGRU   & 27,649 & \textbf{2.193} & \textbf{11.856} & $+$0.098 & \textbf{3.442} \\
            & GRUReLU & 13,825 & 2.366 & 13.501 & \textbf{$+$0.008} & 3.674 \\
            & TCN     & \textbf{6,785}  & 3.706 & 25.422 & $-$0.437 & 5.023 \\
        \hline
    \end{tabular}
\end{table}

\noindent\textbf{Note on parameter counts.} The parameter counts in Table~\ref{tab:application_results} are substantially larger than the benchmark values in Table~\ref{tab:global_ranking} for two reasons: (i)~the input dimension increases from 1 (scalar benchmark sequences) to 5 (joint-angle channels), and (ii)~the hidden dimensions are enlarged to match the added complexity of real-world multivariate data (e.g., hidden size 16~$\to$~64 for GRU-based models). Together, these two factors account for the approximately 14$\times$ increase in parameter count. Counts also differ slightly between interpretations for the same architecture because the output-layer normalization varies per target variable.

\subsection{Interpretation of Error Metrics}

\begin{figure}[!htbp]
    \centering
    \includegraphics[width=\columnwidth]{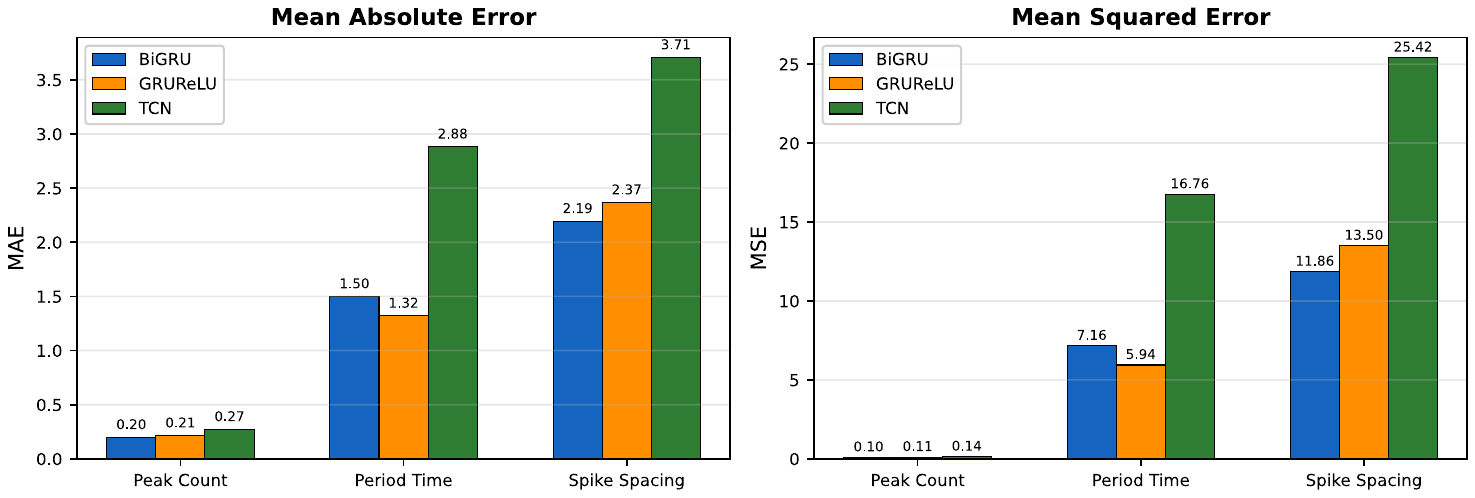}
    \caption{MAE and MSE for the three selected architectures across the three speed interpretations.}
    \Description{Grouped bar chart comparing the MAE of BiGRU, GRUReLU, and TCN across Peak Count, Period Time, and Spike Spacing interpretations.}
    \label{fig:application_mae}
\end{figure}

The practical significance of these error values depends on the output range of each interpretation (Figure~\ref{fig:application_mae}):

\begin{itemize}
    \item \textbf{Peak Count:} The target range is $[0, 4]$. The best model (BiGRU) achieves MAE = 0.198, meaning the predicted peak count deviates from the true value by $\sim$0.2 on average---a relative error of approximately \textbf{5.0\%}. All three models perform well on this interpretation.

    \item \textbf{Period Time:} The target range is approximately $[15, 48]$ frames (range $\approx 33$). The best model (GRUReLU) achieves MAE = 1.320, corresponding to a relative error of approximately \textbf{4.0\%} with respect to the range. This means the estimated cycle duration is off by about 1.3 frames on average.

    \item \textbf{Spike Spacing:} The target range is approximately $[3, 30]$ frames (range $\approx 27$). The best model (BiGRU) achieves MAE = 2.193, corresponding to a relative error of approximately \textbf{8.1\%}. This interpretation is the most challenging because the model must estimate the \emph{average} of multiple inter-cycle distances rather than a single measurement.
\end{itemize}

Overall, the \textbf{Peak Count interpretation yields the best results}, with all three models achieving practically useful accuracy. BiGRU is the strongest performer overall, achieving the best MAE on two of three interpretations, while GRUReLU offers a good accuracy-to-parameter-count trade-off (roughly half the parameters of BiGRU).

TCN, despite being jointly ranked first in the global benchmark, performs weakest on all three real-data interpretations---a reversal that is not coincidental. In the benchmark, TCN excels on scalar ($d=1$) sequences; in the application, the input has five joint-angle channels ($d=5$). TCN's dilated convolutions must now share their capacity across multiple input channels within the same architectural structure, reducing their advantage for detecting 1D temporal patterns. Recurrent models (BiGRU, GRUReLU) integrate multi-channel information naturally through hidden-state updates and are therefore more robust when input dimensionality increases. This highlights an important practical caveat: benchmark rankings on scalar sequences do not automatically guarantee proportional advantage on multivariate real-world inputs.

\subsection{Error Distribution and Real-World Deployment}

\begin{figure}[!htbp]
    \centering
    \includegraphics[width=\columnwidth]{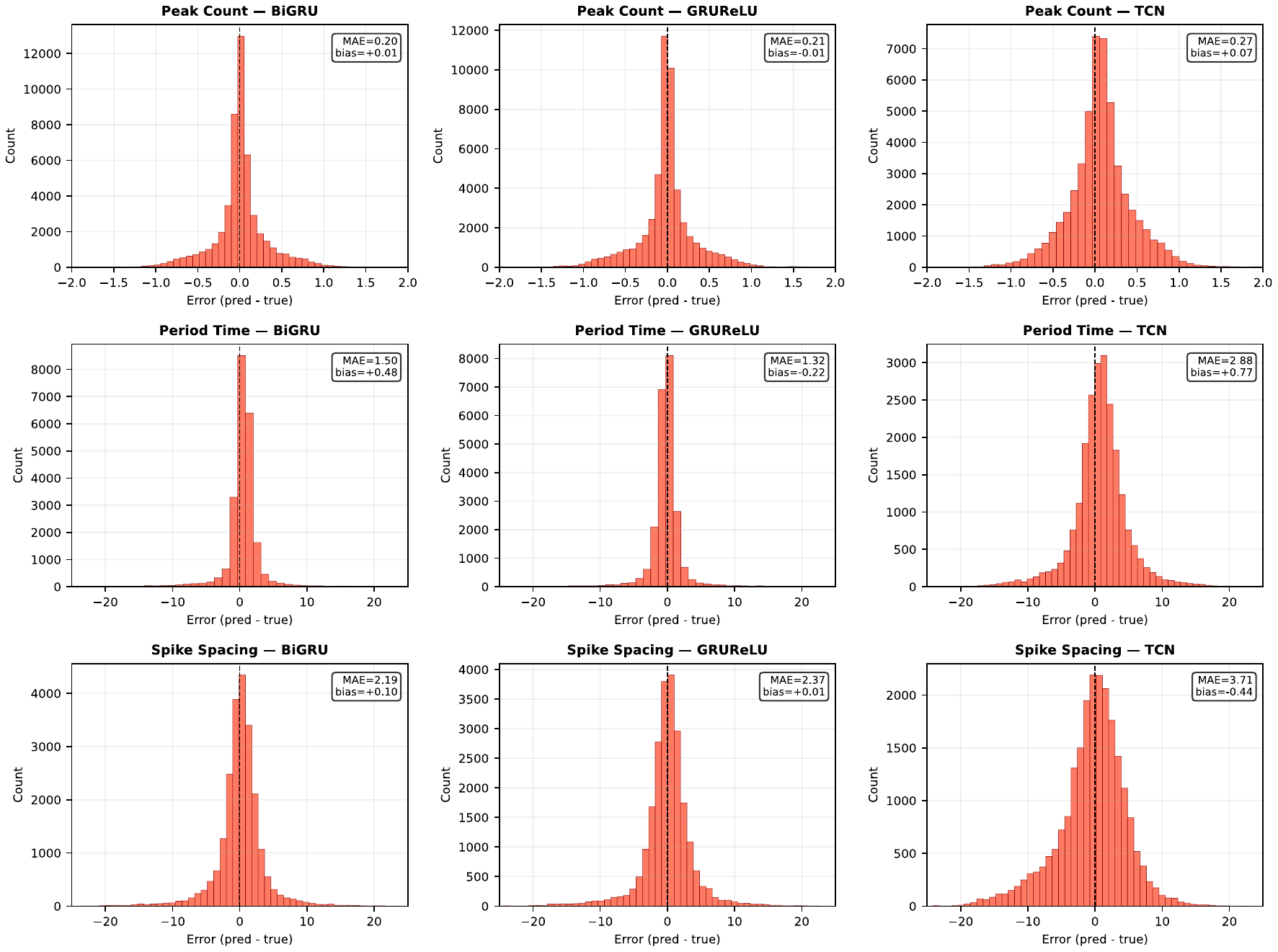}
    \caption{Prediction error distributions ($\hat{y} - y$) for each architecture--interpretation pair (rows: Peak Count, Period Time, Spike Spacing; columns: BiGRU, GRUReLU, TCN).}
    \Description{A 3 by 3 grid of histograms showing the distribution of prediction errors for each combination of speed interpretation (rows) and architecture (columns).}
    \label{fig:application_error_stats}
\end{figure}

Beyond average-case metrics, we evaluate the full \emph{error distribution} to assess deployment suitability: a low mean error is insufficient if individual predictions occasionally deviate wildly, as such outliers can trigger unexpected robotic behavior. We therefore report both signed mean error (Bias) and standard deviation (Std.) for each model--interpretation pair (Table~\ref{tab:application_results}), and inspect the full error histograms (Figure~\ref{fig:application_error_stats}).

Here we distinguish two related but separate quantities: the \emph{error distribution} (the distribution of $\hat{y} - y$, i.e.\ how much each prediction deviates from the true value), and the \emph{spread of predictions around the true value} (captured by the standard deviation of that error). The former reveals systematic bias; the latter determines whether the model is consistently reliable or occasionally catastrophically wrong.

All three models exhibit near-zero bias on the Peak Count interpretation (absolute value below 0.07), confirming that systematic over- and under-prediction cancel out. More importantly, the error standard deviations remain tightly bounded: BiGRU achieves $\sigma = 0.311$ peaks on Peak Count, meaning 95\% of predictions land within $\pm 0.62$ of the true value. The histograms confirm the absence of heavy outlier tails across all interpretations---precisely the property required for reliable real-time deployment.

\section{Conclusions and Future Work}
\label{s:conclusions}

This paper set out to answer a concrete question: \emph{which neural network architectures are best suited for recognizing abstract temporal patterns in sequential data?} Through a systematic benchmark of eighteen architectures across ten tasks---five permutation-invariant and five order-dependent---evaluated under strictly identical conditions with more than 250 experimental configurations, we have arrived at a clear answer.

Four architectures---\textbf{BiGRU}, \textbf{TCN}, \textbf{Conv1D}, and \textbf{GRUReLU}---consistently outperform the field across both task types, occupying the top three places in the global ranking (BiGRU and TCN jointly ranked first with $\Sigma_{\text{rank}} = 31$, Conv1D ranked second with $43$, and GRUReLU ranked third with $46$). Crucially, all four are compact (under 2\,000 parameters in the benchmark configuration), making them directly suitable for real-time deployment in resource-constrained robotic systems.

The practical value of these findings was validated through a robotics-relevant application: estimating the execution speed of dynamic arm gestures from skeletal keypoint data recorded via OpenPose. Three speed interpretations---peak count, period time, and spike spacing---were evaluated on a custom dataset of eight traffic-related gesture classes totaling 256\,710 frames. The results demonstrate that neural networks can estimate gesture speed from skeletal data with high accuracy:

\begin{itemize}
    \item Peak count: MAE = 0.198 (approximately 5\% relative error) with BiGRU
    \item Period time: MAE = 1.320 (approximately 4\% relative error) with GRUReLU
    \item Spike spacing: MAE = 2.193 (approximately 8\% relative error) with BiGRU
\end{itemize}

These error rates are well within the tolerance required for distinguishing fast from slow gesture execution in real-time robotic control, confirming that the benchmark methodology identifies architectures that generalize to real-world data.

\subsection{Future Directions}

Several promising directions remain for future work:

\begin{itemize}
    \item \textbf{End-to-end speed-aware gesture control:} Joint modeling of gesture type and speed could enable robot responses based on both \emph{what} gesture is performed and \emph{how fast}---e.g., fast circular motion triggers emergency stop, slow motion a gradual turn.

    \item \textbf{Gesture-agnostic speed estimation:} A single model that estimates speed independently of gesture identity would improve scalability by removing the requirement for known gesture types at inference time.

    \item \textbf{Noise robustness:} Extending the benchmark with keypoint jitter, occlusions, and natural variability would systematically evaluate architecture robustness in real-world deployment conditions.
\end{itemize}

\begin{acks}
Supported by the \texttt{EKÖP-25-1-I-ELTE-872} project of the EKÖP-25 University Excellence Scholarship Program of the Ministry for Culture and Innovation from the source of the National Research, Development and Innovation Fund.

\end{acks}

\bibliographystyle{ACM-Reference-Format}
\bibliography{introb26}

\end{document}